%% file: Arxiv.tex
\documentclass{article}
\usepackage{arxiv}

\usepackage[utf8]{inputenc} % allow utf-8 input
\usepackage[T1]{fontenc}    % use 8-bit T1 fonts
\usepackage{hyperref}       % hyperlinks
\usepackage{booktabs}       % professional-quality tables
\usepackage{amsfonts}       % blackboard math symbols
\usepackage{nicefrac}       % compact symbols for 1/2, etc.
\usepackage{microtype}      % microtypography
\usepackage{lipsum}		% Can be removed after putting your text content
\usepackage{graphicx}
\usepackage{natbib}
\usepackage{doi}
\usepackage{times}  % DO NOT CHANGE THIS
\usepackage{helvet}  % DO NOT CHANGE THIS
\usepackage{courier}  % DO NOT CHANGE THIS
\usepackage{natbib}  % DO NOT CHANGE THIS AND DO NOT ADD ANY OPTIONS TO IT
\usepackage{caption} % DO NOT CHANGE THIS AND DO NOT ADD ANY OPTIONS TO IT
\usepackage{booktabs}     % for \toprule, \midrule, \bottomrule
\usepackage{multirow}     % for \multirow
\usepackage{threeparttable} % for threeparttable environment
\input{math_commands}

\usepackage{hyperref}
\usepackage{url}
\usepackage{graphicx}
\usepackage{amsmath}
\usepackage{soul}
\usepackage{float}
\usepackage{xcolor}        
\usepackage{algorithm}
\usepackage{algpseudocode}
\usepackage{listings} 
\usepackage{titlesec}
\usepackage{booktabs}
\usepackage{footnote}
\makesavenoteenv{table}
\colorlet{punct}{red!60!black}
\definecolor{background}{HTML}{EEEEEE}
\definecolor{delim}{RGB}{20,105,176}
\colorlet{numb}{magenta!60!black}
\lstdefinelanguage{json}{
    basicstyle=\normalfont\ttfamily\footnotesize,
    numbers=left,
    numberstyle=\scriptsize,
    stepnumber=1,
    numbersep=8pt,
    showstringspaces=false,
    breaklines=true,
    frame=single,
    backgroundcolor=\color{background},
    literate=
     *{0}{{{\color{numb}0}}}{1}
      {1}{{{\color{numb}1}}}{1}
      {2}{{{\color{numb}2}}}{1}
      {3}{{{\color{numb}3}}}{1}
      {4}{{{\color{numb}4}}}{1}
      {5}{{{\color{numb}5}}}{1}
      {6}{{{\color{numb}6}}}{1}
      {7}{{{\color{numb}7}}}{1}
      {8}{{{\color{numb}8}}}{1}
      {9}{{{\color{numb}9}}}{1}
      {:}{{{\color{punct}{:}}}}{1}
      {,}{{{\color{punct}{,}}}}{1}
      {\{}{{{\color{delim}{\{}}}}{1}
      {\}}{{{\color{delim}{\}}}}}{1}
      {[}{{{\color{delim}{[}}}}{1}
      {]}{{{\color{delim}{]}}}}{1},
}

\usepackage[utf8]{inputenc} % allow utf-8 input
\usepackage[T1]{fontenc}    % use 8-bit T1 fonts
\usepackage{hyperref}       % hyperlinks
\usepackage{url}            % simple URL typesetting
\usepackage{booktabs}       % professional-quality tables
\usepackage{amsfonts}       % blackboard math symbols
\usepackage{nicefrac}       % compact symbols for 1/2, etc.
\usepackage{microtype}      % microtypography
\usepackage{xcolor}         % colors

\title{LLM enhanced graph inference for long-term disease progression modelling }

\author{%
Tiantian He, An Zhao, Elinor Thompson, Anna Schroder, Ahmed Abdulaal,\\ Frederik Barkhof, Daniel C. Alexander \\
    Department of Computer Science\\
    University College London\\
    \texttt{tiantian.he.20@ucl.ac.uk}
}

\begin{document}

\maketitle

\begin{abstract}

Understanding the interactions between biomarkers among brain regions during neurodegenerative disease is essential for unravelling the mechanisms underlying disease progression. For example,  pathophysiological models of Alzheimer's Disease (AD) typically describe how variables, such as regional levels of toxic proteins, interact spatiotemporally within a dynamical system driven by an underlying biological substrate, often based on brain connectivity. However, current methods grossly oversimplify the complex relationship between brain connectivity by assuming a single-modality brain connectome as the disease-spreading substrate. This leads to inaccurate predictions of pathology spread, especially during the long-term progression period. Meanhwile, other methods of learning such a graph in a purely data-driven way face the identifiability issue due to lack of proper constraint. We thus present a novel framework that uses Large Language Models (LLMs) as expert guides on the interaction of regional variables to enhance learning of disease progression from irregularly sampled longitudinal patient data. By leveraging LLMs' ability to synthesize multi-modal relationships and incorporate diverse disease-driving mechanisms, our method simultaneously optimizes 1) the construction of long-term disease trajectories from individual-level observations and 2) the biologically-constrained graph structure that captures interactions among brain regions with better identifiability. We demonstrate the new approach by estimating the pathology propagation using tau-PET imaging data from an Alzheimer's disease cohort. The new framework demonstrates superior prediction accuracy and interpretability compared to traditional approaches while revealing additional disease-driving factors beyond conventional connectivity measures.

\textbf{keywords} LLM, spatio-temporal modelling, disease progression

\end{abstract}

\section{Introduction}

Neurodegenerative diseases are characterized by a progressive spread of pathology throughout the brain \cite{Busche2020SynergyDisease}. Elucidating the mechanisms that drive this long-term progression is essential for developing effective treatments. Such pathological propagation is related to brain connectivity, as observed by e.g. Heeley et al. (2009), but that the precise relationship to brain connectivity remains poorly understood. A widely adopted approach to simulating the spread of pathology over the brain during neurodegenerative disease is offered by network diffusion models (NDMs) \cite{zhou2012predicting,raj2012network,seguin2023brain,garbarino2019differences}, which rely on two key components: (i) a graph (often derived from estimates of brain connectivity) that encodes how pathology in one brain region influences others, and (ii) a propagation mechanism that describes how pathology spreads along these connections. For example, \cite{raj2012network,weickenmeier2018multiphysics} assume that pathology dissemination follows structural brain connections measured from MRI. However, using such connectivity measures as direct proxies for graph link strength oversimplifies the complex relationship between disease pathophysiology and brain connectivity. Recent reviews \cite{Young2024Data-drivenBox, Vogel2023Connectome-basedInsight} have identified connectome accuracy as a critical bottleneck in neurodegenerative disease modeling, particularly when capturing the intricate interactions among multiple biological mechanisms. Although several approaches \cite{garbarino2019differences,he2023coupled,thompson2024combining} have attempted to integrate multiple propagation mechanisms or multiple graphs, they typically rely on simple linear combinations that may not capture the full complexity of these interactions.

In contrast to the above methods using fixed graphs, data-driven graph learning methods infer relationships between variables directly from time series data. For instance, \cite{bellot2021neural} et al. introduced a score-based learning algorithm using penalized Neural ODEs to extract relationships from irregular longitudinal data. Despite their potential, these methods often struggle with challenges related to graph identifiability and interpretability, especially when applied to high-dimensional datasets without appropriate clinical constraints.

LLMs offer a novel approach to graph learning by leveraging metadata associated with variables rather than raw data values, enabling inference akin to human domain experts. Prior studies \cite{kiciman2023causal,choi2022lmpriors,long2023can,abdulaal2023causal} have demonstrated their potential in refining graph discovery, integrating multi-modal brain data, and enhancing interpretability. Unlike traditional knowledge graphs, LLMs provide flexible reasoning capabilities and can dynamically incorporate emerging research evidence. This advancement enables the integration of multiple biological modalities, addressing key limitations of pathophysiological and data-driven models in neurodegenerative disease modeling.

A further challenge in modelling disease progression arises from real-world data acquisition constraints. Medical imaging data are often collected irregularly and over limited time frames, making it hard to estimate the long-term population-level trajectories across decades of progression from such sparse and irregularly sampled patient data sets. Traditional long-term longitudinal analysis methods typically assume that observation timestamps are known. This scenario creates a unique challenge: the simultaneous optimization of both the temporal placement of observations and the inference of the underlying relationships that drive disease trajectories.

In this work, we address these two challenges by developing a novel LLM-guided disease progression model that simultaneously reconstructs the temporal evolution and spatial interactions of regional brain pathology markers from irregular spatio-temporal data. Unlike conventional methods that rely on predetermined timestamps, our framework jointly optimizes the temporal positioning of observations while uncovering the network of pathological interactions that govern disease progression constrained by multi-modal knowledge from LLM.

Our key contributions are:

\begin{itemize}
    \item We propose a framework to construct a long-term continuous disease progression trajectory from irregular snapshots, while simultaneously performing graph learning for the resulting series. This framework can serve as a disease staging system for patients.
    \item We propose a way of embedding multi-modal expertise knowledge derived from LLMs into graph learning as a substrate for disease progression to achieve higher prediction accuracy and improved identifiability with a sparser graph. Compared with biological graphs derived from imaging, the LLM-based approach removes redundant links and suggests missing links supported by recent literature.
\end{itemize}

\section{Methodology} 

\subsection{Problem Statement}

The method assumes the existence of a dataset containing data from one or more exams of N patients.  Each exam acquires D measurements, or "biomarkers".  Thus for each subject $i = 1, ..., N$ the observation from exam $j = 1, ... J_i$, is a set $C_i = \{c_{i,j}^d\}$ of scalar biomarkers. The algorithm aims to:
\begin{itemize}
    \item Construct the long-term cohort-level disease progression trajectory, starting from the very early pathology onset time to the late disease stage, from the snapshots of individual-level observations by allocating subjects on the temporal axis.
    \item Identify a $D \times D$ graph $\mathbf{G}$ that encapsulates the influence of each biomarker on every other. A graph element $\mathbf{G}_{k,p}$ indicates the extent to which a biomarker $k$ is likely to affect the biomarker $p$. Thus the trajectory is dependent on the underlying graph, as ${\mathbf{c}}(t) = f_\mathbf{G}(t)$.
    \item Dynamically update the relative location of each individual on the cohort-level trajectory and the trajectory itself.
    
    %We consider this graph identification in two distinct scenarios, namely: 
    %\begin{itemize}
        %\item The graph estimation for pathophysiological models where the structure of $f_\mathbf{G}(t)$ is predefined according to specific physical processes, thus only the graph for how regional biomarkers interact with each other needs to be optimized;
        %\item The graph estimation for purely data-driven graph learning algorithms where both the graph $\mathbf{G}$ and the structure of $f_\mathbf{G}(t)$ are unknown.
    %\end{itemize}
\end{itemize}

\subsection{Overview of the proposed framework}
\begin{figure}%[H]
\centering
\includegraphics[scale = 0.47]{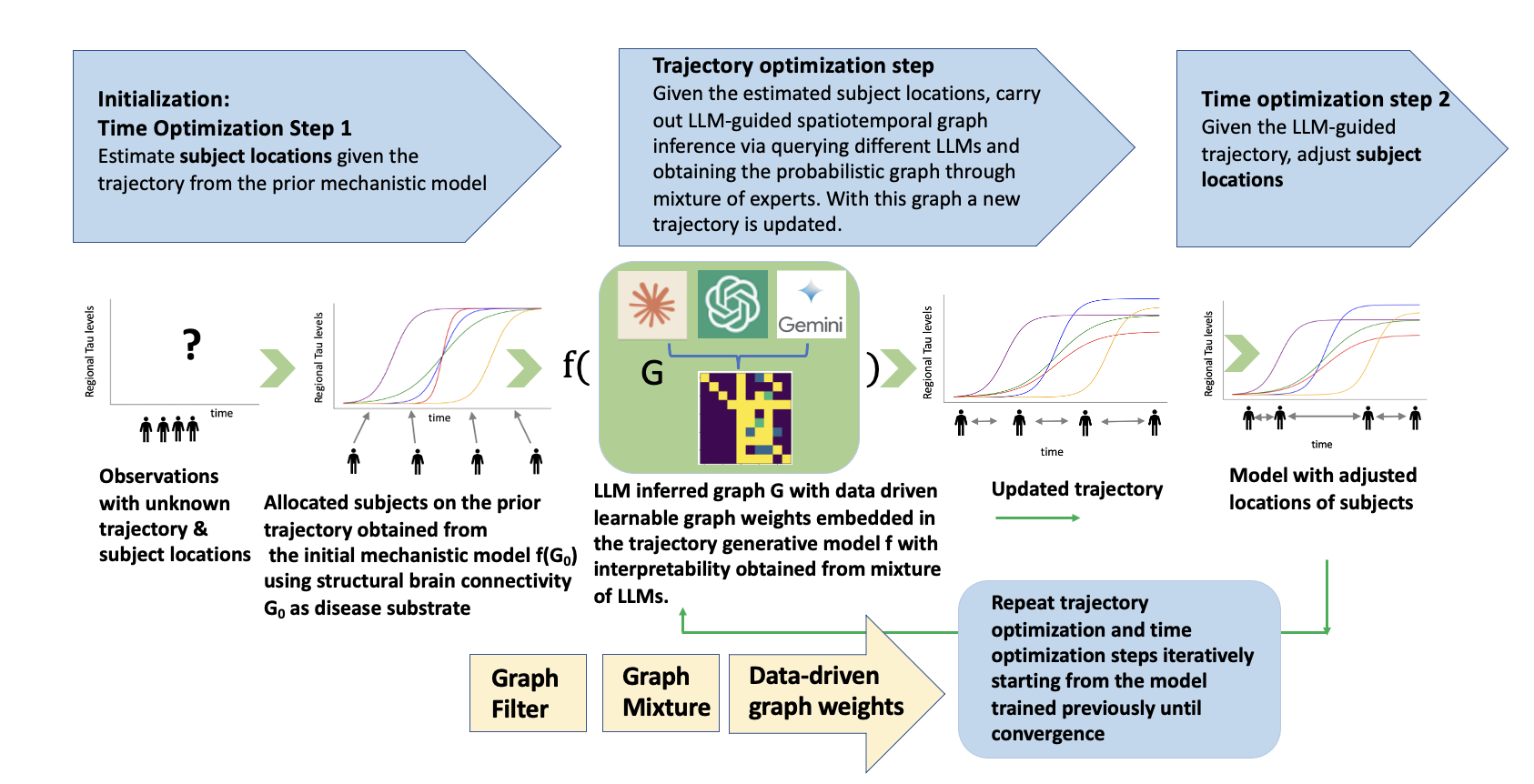}
\caption[Model Overview]{\textbf{Model Overview} The proposed framework for constructing a full disease progression process from snapshots, by iteratively estimating subject locations and the embedded graph. The graph plays a dominant role in shaping the disease trajectory. Graph inference includes LLM query, graph filtering and data-driven graph weights learning.} 
\label{framework}
\end{figure}

Figure \ref{framework} shows the overview of the framework. 
We jointly reconstruct the trajectory of disease biomarkers and obtain the relative locations of each individual on the progression time axis through the \textbf{dual optimization} method, as proposed in \cite{he2025stage}.The key of the trajectory update is graph learning, which includes several steps: first, we obtain the initial graph $\mathbf{G}$ through querying the large language model (LLM);  then we process $\mathbf{G}$ by graph filtering and finally refine the weights of the non-zero elements of the graph in a data-driven way. We then input the graph $\mathbf{G}$ into the generative model to estimate $f_\mathbf{G}(t)$.

\subsection{LLM-guided graph inference}
\label{graph_section}
\subsubsection{Querying a probabilistic graph from LLMs}

%This section needs a rethink.  I would turn it on its head.  Separate sections on a) specific LLM problem statement - what are we trying to get from them, b) prompt design; c) the choice of LLMs themselves; d) query strategy - how exactly we obtain the numerical graph elements from the LLMs.

To address the identifiability challenge in graph learning, improve robustness, and enhance interpretability, we propose a novel strategy that leverages large language models (LLMs) to infer disease-relevant brain graphs. Unlike traditional data-driven approaches which often lack stability and require large datasets, or simple linear combinations of existing graphs, LLMs offer several distinct advantages:

1. Enhanced Flexibility: Users can easily test hypotheses about pathophysiological factors by adjusting prompts, rather than being constrained by existing databases. LLMs can explore factors beyond current data availability, such as genetic influences, enabling rapid hypothesis testing.

2. Complex Integration: LLMs can capture more sophisticated relationships than linear combinations of different factors, providing a richer understanding of brain connectivity patterns.

3. Reasoning Capability: Beyond identifying connections, our system provides pathophysiological explanations for each relationship, grounding each link with reasoning and thus improving identifiability.

4. Adaptability: LLMs rapidly incorporate new research findings and adjust relationships based on emerging evidence which has the potential to improve the false positive or negative errors in the measured biological graphs.

\subsubsection{Prompt Design: Structuring Inputs for Reliable Graph Inference}

Effective LLM-based graph inference requires carefully structured prompts that encapsulate essential neurobiological factors underlying disease propagation. Our prompt architecture consists of five key components:

1. \textbf{Context Setting}: Establishes the expert role of LLM.

2. \textbf{Task Framing}: Explains the relationship between brain graphs and disease patterns, focusing on tau pathology in Alzheimer's disease. For each region of interest (ROI), the model infers connections to all other regions based on predefined components.

3. \textbf{Anatomical Framework}: Defines brain regions using the Desikan-Killiany Atlas and clarifies hemisphere nomenclature.

4. \textbf{Biological Factors}: Incorporates five primary factors that may influence pathology spread \cite{thompson2024combining}, including: structural connectivity, Cortical morphology similarity, Microstructural profile covariance, Geodesic proximity, and Functional connectivity. We aim to compare the mixture of LLM with a linear combination of these factore to see if LLM can improve model accuracy. Additionally, we explore an extended 7-factor prompt that includes neurotransmitter density \cite{soskic2024effects} and metabolic correlation maps \cite{adams2019relationships} that we don't have in the database.

5. \textbf{Output Structuring}: Defines a standardized response format where the LLM assigns connection strengths (0-1) for each potential edge and provides reasoning documentation.

To ensure reliable graph inference, we employ multiple state-of-the-art LLMs, including Claude-3.5-Sonnet, GPT-4-turbo, and Gemini 1.5. Each LLM is queried three times with a temperature of 0.25, and responses are averaged to obtain the final probabilistic graph.

\subsubsection{Graph Filtering}

To avoid identifiability and overfitting problems, we apply thresholding to the LLM graphs by only keeping edges with weights above a specific threshold. In this way we obtain a sparse binary graph, where the weight of each non-zero element can be then learnt in the disease progression task.

%\begin{figure}%[H]
%\includegraphics[scale = 0.45]
%{plots/filtered_graph.png}
%\centering
%\caption[Filtered Graph]{\textbf{Graph Filtering}. The plot shows the process of filtering graphs by increasing the threshold. } 
%\label{traj_visuals}
%\end{figure}

\subsection{Embedding the graph into the dynamical system for trajectory construction}
\label{pathophysiological_section}

Now we embed the graph $\mathbf{G}$ into the generative model $f_\mathbf{G}(t)$ and compare this model's output with actual observations to determine the prediction error. Through this process, the weight of $\mathbf{G}$ is shaped by integrating both the expert insights from LLM and the capabilities of data-driven analysis. 

\textbf{Model 1 (baseline) - Spreading model for regional tau on the weighted structural connectome}
The baseline pathophysiological model for tau propagation consists of the diffusion process along the structural connectome and the local production of the pathology:
\begin{equation}
\label{eq:baseline NDM}
\frac{d \mathbf{c}}{d t}=-k [\mathbf{L} \mathbf{c}(t)]+\alpha \mathbf{c}(t) \odot[v  \mathbf{p} -\mathbf{c}(t)]
\end{equation}
where $\mathbf{L_{bio1}}$ is the normalized Laplacian of the structural connectivity matrix obtained via $\mathbf{L_{bio1}} = \mathbf{D_{bio{1}}} - \mathbf{A_{bio1}}$, where the diagonal degree matrix $\mathbf{D_{bio1}}$ is the row sum of the weights of the edges connected to each vertex; and the weighted adjacent matrix $\mathbf{A_{bio1}}$ is the filtered binary structural connectome $\mathbf{G_{bio1}}$ multiplied by the learnt weight $\mathbf{W_{bio1}}$ from the data representing the extent of regional interactions, resulting in $\mathbf{A_{bio1}} = \mathbf{G_{bio1}} \odot \mathbf{W_{bio1}}$. Parameters $k$, $\alpha$ and $v$ represent pathology spreading rate, aggregation rate, and convergence level, respectively. The carrying capacity $\mathbf{p}$ is calculated from the 99th percentile of regional tau distribution. 

\textbf{Model 2 (baseline) - Coupled-mechanisms via multiple connectomes}
Building upon Model 1, the linear combination of multiple connectomes can be included, according to \cite{thompson2024combining}:
\begin{equation}
\label{eq:mix-NDM}
\frac{d \mathbf{c}}{d t}=-k [(w_1\mathbf{L_{bio1}} + w_2\mathbf{L_{bio2}} + w_3\mathbf{L_{bio3}} + w_4\mathbf{L_{bio4}} + w_5\mathbf{L_{bio5}}) \mathbf{c}(t)]+\alpha \mathbf{c}(t) \odot[v \mathbf{p}-\mathbf{c}(t)]
\end{equation}
where $\mathbf{L_{bio1}}$ through $\mathbf{L_{bio5}}$ represent Laplacians derived from structural, functional, morphological, geodesic, and microstructural connectomes, respectively obtained using the same calculation as before, but the mixture is limited to simple linear combinations.

\textbf{Model 3 (proposed) - Coupled-mechanisms via LLM-informed connectome integration}
Our proposed model enhances connectivity integration through LLM-derived relationships:
\begin{equation}
\label{eq:llm-NDM}
\frac{d \mathbf{c}}{d t}=-k [\mathbf{L_{LLM}} \mathbf{c}(t)]+\alpha \mathbf{c}(t) \odot[v\mathbf{p}-\mathbf{c}(t)]
\end{equation}
where $\mathbf{L_{LLM}} = \mathbf{D_{LLM}} - \mathbf{A_{LLM}}$ and $\mathbf{A_{LLM}} = \mathbf{G_{LLM}} \odot \mathbf{W}$. The probabilistic graph $\mathbf{G_{LLM}}$ is queried from LLMs (Claude3.5, GPT4-turbo, Google Gemini 1.5 Pro) using prompts that consider multiple biological factors beyond structural connections. 

\section{Data}
\subsection{Data Processing and Subject Selection}
We analyze tau-PET SUVRs from the Alzheimer’s Disease Neuroimaging Initiative (ADNI) database (adni.loni.usc.edu), excluding subcortical regions due to off-target binding effects. Our cohort consists of 255 amyloid-positive subjects with 378 total observations (1-4 scans per subject). Subjects are included if they show tau-positivity in at least one region, determined using a two-component Gaussian mixture model. Tau values are normalized to [0,1] across all participants and regions. For pathophysiological models, we initialize from bilateral inferior temporal cortex regions, while data-driven methods use earliest disease stage observations.
\subsection{Training Protocol}
We implement 3-fold cross-validation with 35 subjects each in validation and test sets, maintaining all longitudinal scans from individual subjects within their assigned sets. In the training process, the model constructs a disease progression trajectory. The validation step happens after an epoch of trajectory optimization on the training data, i.e. the subjects from the validation set are allocated on the trajectory from each training epoch through stage optimization. For new subjects, the model can determine their relative disease stage through time optimization and provide interpretable insights about regional interactions through the LLM-guided graph.

\section{Experiments and Results}

In this section, we present a comprehensive evaluation of our approach including four parts:

We start with the motivation experiment showing that without proper physical constraint, traditional graph learning methods provide different graphs with similar loss. This illustrates the need for expert knowledge.  

Then, we demonstrate that the LLM-guided graph improves prediction accuracy in the disease progression model. We compare the performance of models that embed the LLM graph (\cite{weickenmeier2018multiphysics}) against those using only the structural connectome or a linear combination of brain connectivity modalities (\cite{thompson2024combining}) at different extents of graph sparsity. %To assess the impact of graph sparsity, we gradually increase the threshold applied to the graphs and evaluate prediction performance at different sparsity levels. Our results show that the LLM-guided graph—constrained by coupled mechanisms and refined through thresholding—achieves superior performance while utilizing fewer parameters.

Next, we perform a series of graph analyses to compare the LLM-derived graph with existing biological graphs. Specifically, we:
\begin{enumerate}
    \item Show that the LLM graph exhibits several patterns similar to those observed in established biological graphs, highlighting its capability to merge multi-modal information from different brain connectomes.
    \item Analyze novel links identified by the LLM that rarely appear in traditional biological graphs, and provide supporting clinical evidence from the literature that these links may offer new insights.
    \item Present the reasoning output from the LLM, which further explains the structure of the graph and enhances interpretability by offering a deeper understanding of the underlying mechanisms.
\end{enumerate}

Finally, we conduct an ablation study on different prompt formulations to explore their impact on the performance of the LLM. This study provides further insights into the role of prompt design in optimizing our model's effectiveness.

\subsection{graph learning from time series without physical constraints}

\begin{figure}
\centering
\includegraphics[scale = 0.2]{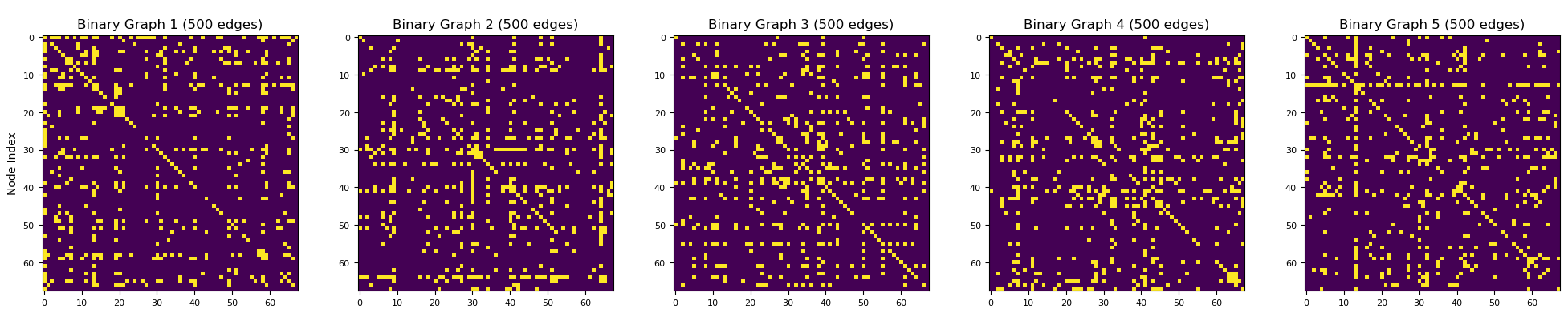}
\caption[Different graph learning outputs from data-driven model NGM ]{Different graph learning outputs from data-driven model NGM}
\label{NGM_plots}
\end{figure}

We first investigate the performance of the Neural Graphical Model (NGM) \cite{bellot2021neural}, a data-driven approach that does not incorporate explicit physical constraints. As shown in Figure \ref{NGM_plots}, two separate runs of the NGM—using identical data and hyperparameters—produce markedly different graph structures. This variability indicates that without appropriate physical constraints, traditional graph learning methods may converge to different graphs despite similar loss values. Such instability underscores the importance of integrating domain expertise and physical constraints into the graph inference process.

\subsection{Comparison of different graphs embedded in pathophysiological models}

We assess tau propagation across brain regions using pathophysiological models. For each graph type, we identify optimal binary graphs through thresholding connectivity measures. Low thresholds have identifiability issues due to redundant pathological propagation paths, while high thresholds maintain key connections but risk failure when critical brain regions disconnect. We target the sparsest graph that accurately captures pathology propagation to prevent overfitting. We define the highest threshold that can retain the performance as the "the critical threshold", and the corresponding minimum number of edges to retain the model performance "the critical edge number", defined as $N^{*}_{edge}$.

\begin{figure}
\centering
\includegraphics[scale = 0.36]{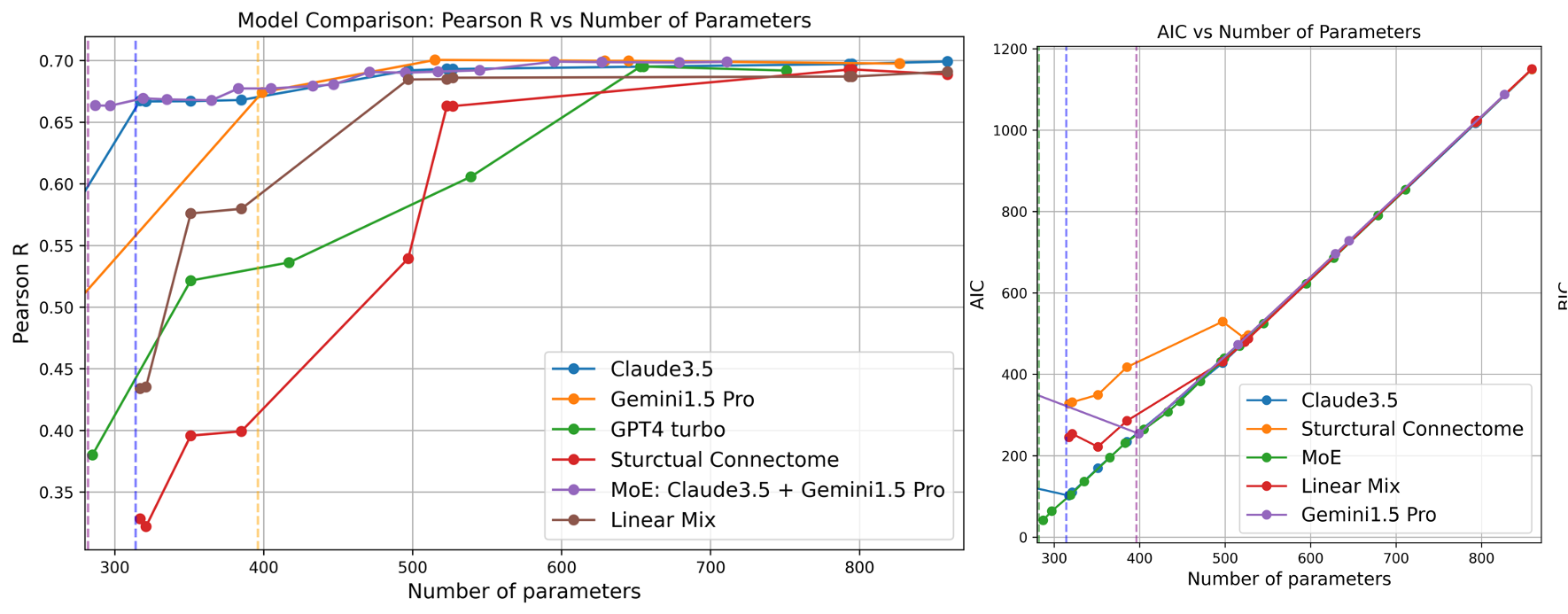}
\caption[Model comparison]{\textbf{Model Performance: R correlation on test set vs parameter number (Left); AIC on training set vs parameter number (Right)}. The dashed vertical lines represent the critical edge numbers of LLMs. The graph obtained from the mixture of LLMs provides the lowest AIC at the smallest parameter number, followed by Claude 3.5. As the number of learnable parameters increases, all models tend to have the same performance level.  The LLM-based graphs allow the model to retain high performance to much greater sparsity levels than the connectivity-based graphs. } 
\label{combined-perfom}
\end{figure}

Table \ref{critical-weighted} shows the LLM-derived graphs achieve better predictions with fewer parameters compared to alternative models at the corresponding the critical threshold.
Figure \ref{combined-perfom} illustrates Pearson R correlation and AIC versus parameter count. Dense graphs exhibit poor identifiability due to similar cross-type performance. The sparse LLM graph demonstrates superior fitting, informing our final data-driven weighted mixed LLM graph. Predicted versus observed tau progression visualization appears in Appendix \ref{brain_map}.

\begin{table}
    \centering
    \caption{\textbf{pathophysiological Model Comparison with different graph embedded}}
    \begin{tabular}{lccccc}
        \toprule
        \textbf{Model Name} &  $\mathbf{N^{*}_{edge}}$ & \textbf{Test SSE} & \textbf{Test Pearson R} & \textbf{Test AIC} \\
        \midrule
        Claude 3.5 Sonnet&  314 & 14.05 ± 1.33 & 0.66 ± 0.03 & 541.49 ± 14.61\\
        Structural Connectome  &  314 & 24.89 ± 2.36 & 0.36 ± 0.03 & 576.96 ± 12.51\\
        \midrule
        Gpt4-turbo & 650 & 13.17 ± 1.65 & 0.70 ± 0.02 & 1209.27 ± 17.19\\
        Structural Connectome & 650 & 13.38 ± 1.86 & 0.68 ± 0.01 & 1210.12 ± 18.12\\
        \midrule
        Gemini Pro 1.5 &  396 & 14.09 ± 1.73 & 0.67 ± 0.02 & 705.47 ± 16.89\\
        Structural Connectome &  396 & 19.10 ± 1.75 & 0.51 ± 0.02 & 724.59 ± 12.85\\
        \midrule
        Mixed LLMs &  \textbf{284} & 14.27 ± 1.41 & 0.64 ± 0.03 & \textbf{482.40 ± 15.21}\\
        Linearly-mixed Connectomes & 314 & 22.50 ± 3.02 & 0.46 ± 0.02 & 570.40 ± 15.61\\
        \bottomrule
    \end{tabular}
    \label{critical-weighted}
\end{table}

\subsection{Graph Analysis}

In addition to predictive performance, we validate the LLM-derived graph by comparing its structural features to established biological graphs and by examining the LLM’s reasoning output.

\subsubsection{Comparison with biological graphs}

\begin{figure}%[H]
\centering
\includegraphics[scale = 0.15]{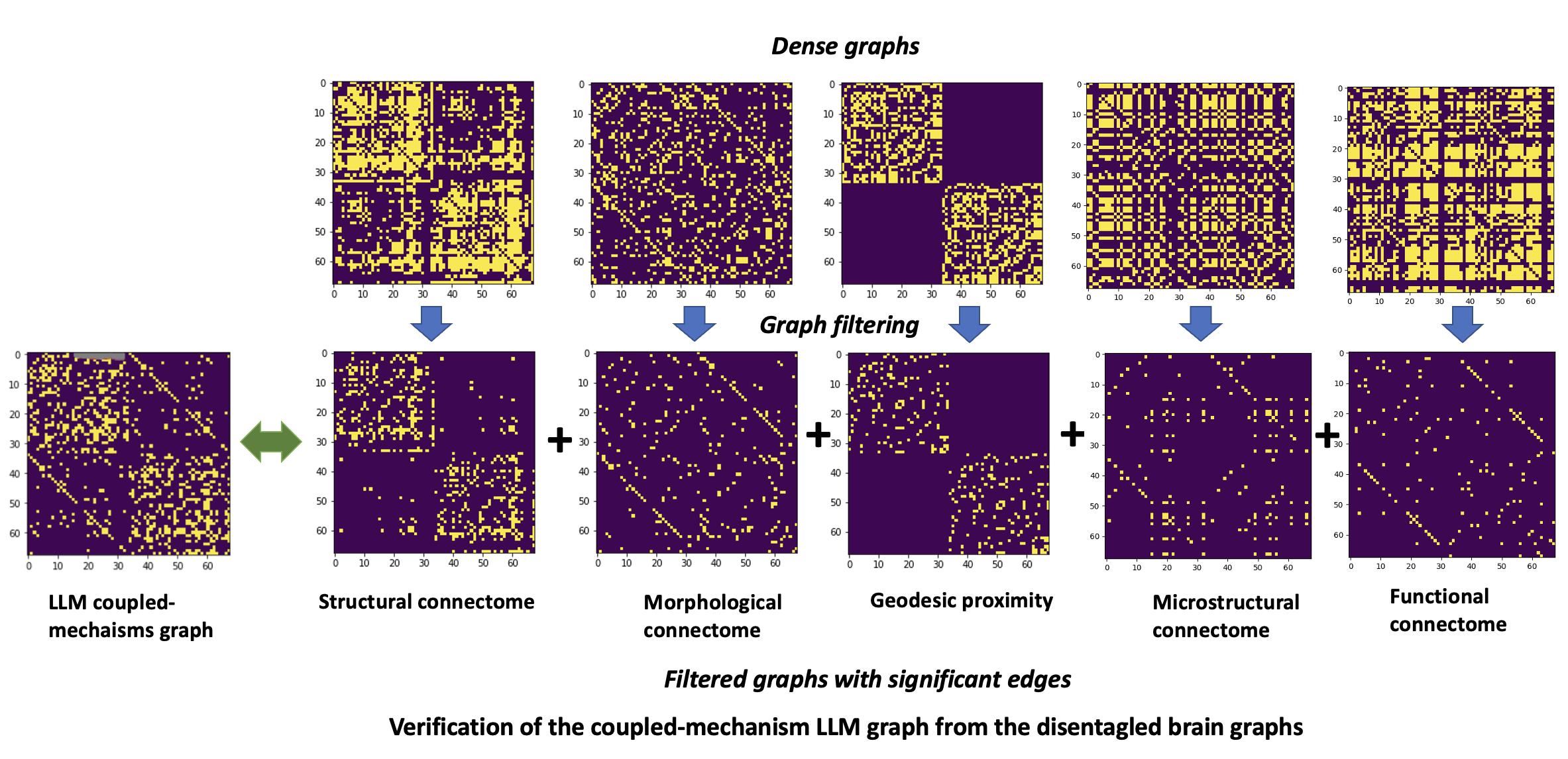}
\caption[Verification of the LLM graph]{\textbf{Verification of the LLM graph} }
\label{disentagle_graph}
\end{figure}

Figure \ref{disentagle_graph} compares the LLM-derived graph with several brain connectivity patterns, including structural, functional, geodesic, and morphological connectomes. In the top row, the dense graphs are shown, while the bottom row displays the corresponding filtered graphs, which highlight significant edges. Key similarities include block-like clusters in top-left and bottom-right regions (matching structural and geodesic patterns), consistent diagonal elements (aligned with functional connectome), etc. Table \ref{tab:global-similarity} quantitatively displays similarity through multiple graph topological analyses. These parallel patterns suggest that LLM mechanisms may mirror fundamental principles of brain connectivity organization across structural, functional, and geodesic dimensions.

\label{graph_verity}

\begin{table}[htbp]
\centering
\caption{Global Similarity Metrics Between Graphs}
\begin{tabular}{lcccc}
\hline
\textbf{Graph Pairs} & \textbf{Frobenius} & \textbf{Pearson} & \textbf{Spearman} & \textbf{Edge Overlap} \\
\hline
Claude-Structure & 0.2445 & 0.5450 & 0.5450 & 0.6007 \\
Claude-Proximity & 0.2747 & 0.5933 & 0.5933 & 0.6431 \\
Claude-Morphology & 0.1535 & 0.3638 & 0.3638 & 0.4417 \\
Claude-Function & 0.1501 & 0.3558 & 0.3558 & 0.4346 \\
Claude-Microstructure & 0.0940 & 0.2068 & 0.2068 & 0.3039 \\
Claude-Random & 0.0335 & 0.0014 & 0.0014 & 0.1237 \\
\hline
\end{tabular}
\label{tab:global-similarity}
\end{table}

%\label{graph_verity}
%\begin{figure}
%\centering
%\includegraphics[scale = 0.22]{plots/graph_hotmap.png}
%caption[Verification of the LLM graph]{\textbf{Verification of the LLM graph} }
%\label{disentagle_graph}
%\end{figure}

We carry out verification of Large Language Model (LLM) coupled-mechanisms through comparison with disentangled brain connectivity patterns on the right, as is shown Figure \ref{disentagle_graph}. 

\subsubsection{link clues from literature}

We then concentrate on the difference between the LLM graph and the biological graph. The connectivity links presented in the table were selected through a systematic frequency-based analysis comparing Large Language Model (LLM) predictions against five distinct biological brain networks. The selection process specifically identified connections that were predicted by the LLM but appeared infrequently across the biological networks, making them particularly interesting cases for investigation. For each connection, we calculated its occurrence frequency across all five biological networks, and prioritized those that were predicted by the LLM but had low representation in the empirical data. This approach helped identify potentially novel or understudied connections that the LLM deemed important, possibly revealing connections that are functionally or pathologically relevant but might be difficult to detect using traditional neuroimaging methods alone. Table \ref{tab:literature} displays the top 5 links that suggested by LLM but rarely appear in biological graphs, together with the literature as the clues of why this might be the case.

\subsubsection{Identifiability from LLM reasoning}

Figure \ref{explains} displays an example of the output from Claude-3.5-Sonnet, when queried about the disease-related interactions between the left rostral anterior cingulate cortex and other 67 cortical brain regions defined in the Desikan-Killiany Atlas. Apart from the five factors we explicitly provide, we further encourage the LLM to think about other factors that might contribute to disease spread (displayed in red). As a result, factors like the effect of neurotransmitter systems are proposed. This has been quantitatively shown to have a significant influence on the spreading of regional tau by \cite{soskic2024effects}. Reasoning from other LLMs can be found in the Appendix.

\begin{figure}%[H]
\centering
\includegraphics[scale = 0.17]{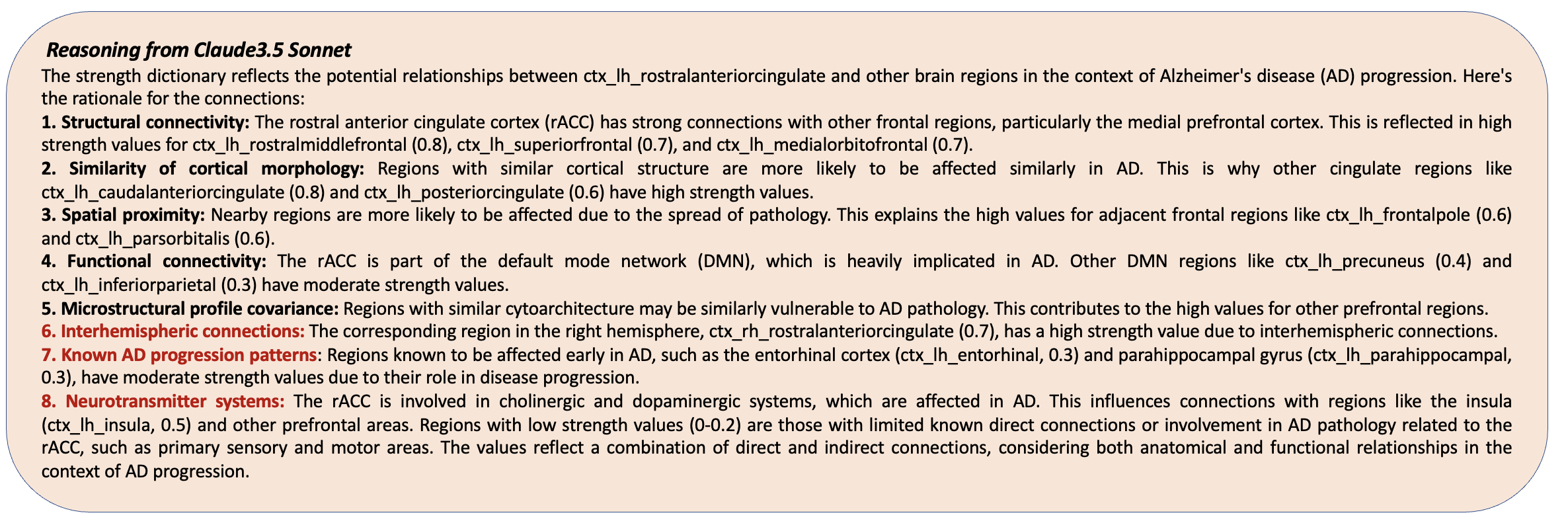}
\caption[One example of reply from Claude 3.5]{This figure displays one representative example of an output from Claude 3.5. Factors in red (6 - 10) are those which weren't mentioned in the prompt.} 
\label{explains}
\end{figure}

\begin{table}[htbp]
\centering
\footnotesize
\begin{tabular}{p{0.3\textwidth}|p{0.25\textwidth}|p{0.35\textwidth}}
\hline
\textbf{Region Pair} & \textbf{Publication} & \textbf{Key Finding} \\
\hline
ctx\_lh\_bankssts $\rightarrow$ ctx\_lh\_superiortemporal & 
Jung et al. (2017) \par Cortex & 
Direct structural connections between bankssts and superior temporal regions support language network integration \\
\hline
ctx\_lh\_caudalanteriorcingulate $\rightarrow$ ctx\_lh\_superiorfrontal & 
Wang et al. (2009) \par Human Brain Mapping & 
Strong structural and functional connectivity between ACC and superior frontal regions in executive control network \\
\hline
ctx\_lh\_caudalanteriorcingulate $\rightarrow$ ctx\_lh\_caudalmiddlefrontal & 
Greicius et al. (2004), Proceedings of the National Academy of Sciences & 
Functional connectivity between the anterior cingulate and frontal regions is altered in Alzheimer's disease, suggesting a potential pathway for disease propagation\\
\hline
ctx\_lh\_frontalpole $\rightarrow$ ctx\_rh\_frontalpole & 
Liu et al (2020) Nat. Commun. & 
The lateral frontal pole (FPl) is involved in regulating action tendencies elicited by emotional stimuli, suggesting interhemispheric connections are important for emotional processing and control. \\
\hline
ctx\_lh\_bankssts $\rightarrow$ ctx\_lh\_middletemporal & 
Petrides Michael (2023) J. Comp. Neurol.  & 
STS integrates multisensory information and connects with the middle temporal gyrus for semantic processing. \\
\hline
\end{tabular}
\caption{Key Brain Region Connections and Their Supporting Evidence}
\label{tab:literature}
\end{table}

\subsection{Ablation study of prompt components}

\begin{figure}
\centering
\includegraphics[scale = 0.28]{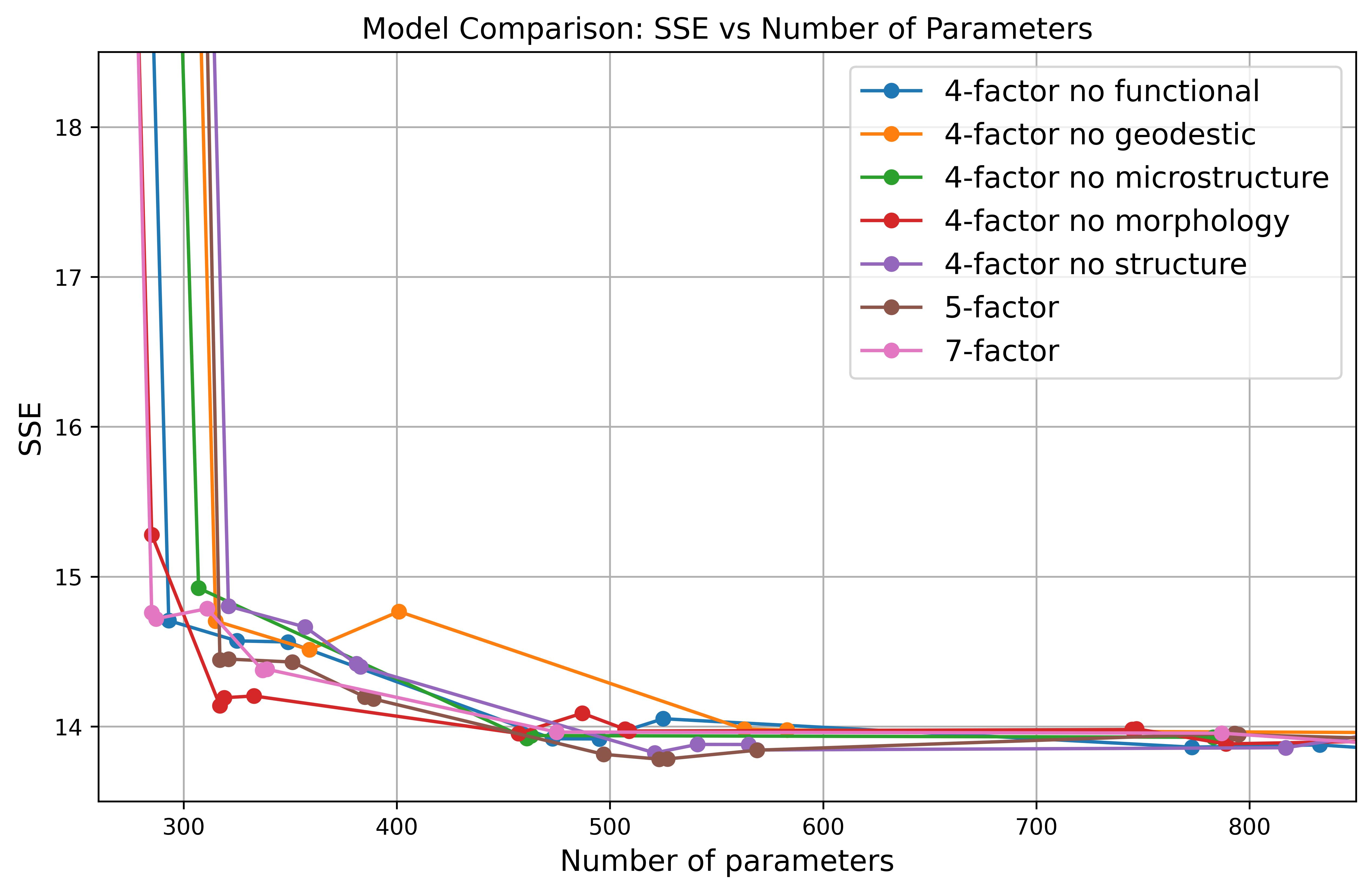}
\caption[Model comparison]{\textbf{Ablation study of different prompts - SSE vs number of remaining edges}.} 
\label{capacity-abulation}
\end{figure}

Figure \ref{capacity-abulation} displays the performance of the best LLM, Claude 3.5, across the different prompts. We consider the original 5-factor prompt; removing each different factor from the original prompt (4-factor prompts); as well as the 7-factor prompt, where two more factors (neurotransmitter density as suggested by \cite{soskic2024effects} and metabolic correlation map as suggested by \cite{adams2019relationships}) have been added. The 5-factor prompt provides the lowest overall test SSE while removing the geodesic proximity significantly decreases the accuracy. The "7-factor" prompt offers a way of extending the knowledge outside the existing five connectomes that are available in the MICA-MICS database by explicitly adding two more features to the prompt. This prompt has further decreased the critical edge number compared with the 5-factor prompt.

% Required packages:
% \usepackage{array}
% \usepackage{amsmath} % For \rightarrow
% Required packages:
% \usepackage{array}

% Required packages:
% \usepackage{array}

\section{Conclusions}
We propose a novel framework designed to construct long-term continuous disease progression trajectories from irregular snapshots while simultaneously performing graph learning on the generated long-term series. By coupling multiple mechanisms from LLMs, our model surpasses the classic pathophysiological model, delivering higher prediction accuracy. Furthermore, by integrating LLMs as constraints in data-driven graph learning methods for time series, our approach not only accelerates and stabilizes convergence but also enhances identifiability and interpretability. For future work, we will look at other indicators of neurodegeneration other than tau. And we will do more exploration to increase LLM performance. This framework can be easily adapted to other domains since the expertise comes from LLM rather than any specific knowledge base.

% References follow the acknowledgments in the camera-ready paper. Use unnumbered first-level heading for
% the references. Any choice of citation style is acceptable as long as you are
% consistent. It is permissible to reduce the font size to \verb+small+ (9 point)
% when listing the references.
% Note that the Reference section does not count towards the page limit.
% \medskip

% {
% \small

% [1] Alexander, J.A.\ \& Mozer, M.C.\ (1995) Template-based algorithms for
% connectionist rule extraction. In G.\ Tesauro, D.S.\ Touretzky and T.K.\ Leen
% (eds.), {\it Advances in Neural Information Processing Systems 7},
% pp.\ 609--616. Cambridge, MA: MIT Press.

% [2] Bower, J.M.\ \& Beeman, D.\ (1995) {\it The Book of GENESIS: Exploring
%   Realistic Neural Models with the GEneral NEural SImulation System.}  New York:
% TELOS/Springer--Verlag.

% [3] Hasselmo, M.E., Schnell, E.\ \& Barkai, E.\ (1995) Dynamics of learning and
% recall at excitatory recurrent synapses and cholinergic modulation in rat
% hippocampal region CA3. {\it Journal of Neuroscience} {\bf 15}(7):5249-5262.
% }

\bibliography{arxiv}
\bibliographystyle{arxiv}

\appendix

\section{Appendix}
\textbf{Summary of Appendices.}
\begin{itemize}
    \item Evaluation of learnt graph on synthetic data
    % \item Prompts
   \item Disease Progression Visualization
   \item More interpretation from LLMs
    \item Related work
\end{itemize}

\subsection{Evaluation of learnt graph on synthetic data}
\label{groundtruth_graph}

Apart from NGM discussed previously, there are other graph learning methods for time series which do not explicitly aim to generate time series as the methods before. Instead, they focus more on discovering the graph of how different variables interact with each other from time series. Thus, they can also be baselines to compare with our proposed method on the accuracy of graph inference. The Structural Vector Autoregression Model (SVAM) (Hyvärinen et al., 2010), an extension of the LiNGAM algorithm to time series, is another representative model. PCMCI (Runge et al., 2017), a representative independence-based approach to structure learning with time series data, extends the PC algorithm. Another method, Dynamic Causal Modelling (DCM), is a representative two-stage collocation approach in which derivatives are first estimated on interpolations of the data, and a penalized neural network is learned to infer G (extending the linear models of Ramsay et al., 2007; Wu et al., 2014; Brunton et al., 2016). However, the optimization goal is to minimize the modeled derivatives with the interpolated derivative of the data, rather than directly optimizing the trajectory itself. Thus, for those methods, we cannot compare the modelling accuracy of disease progression, but instead, we evaluate the obtained graph with the ground truth graph from synthetic data. 

\begin{figure}[H]
\centering
\includegraphics[scale = 0.3]{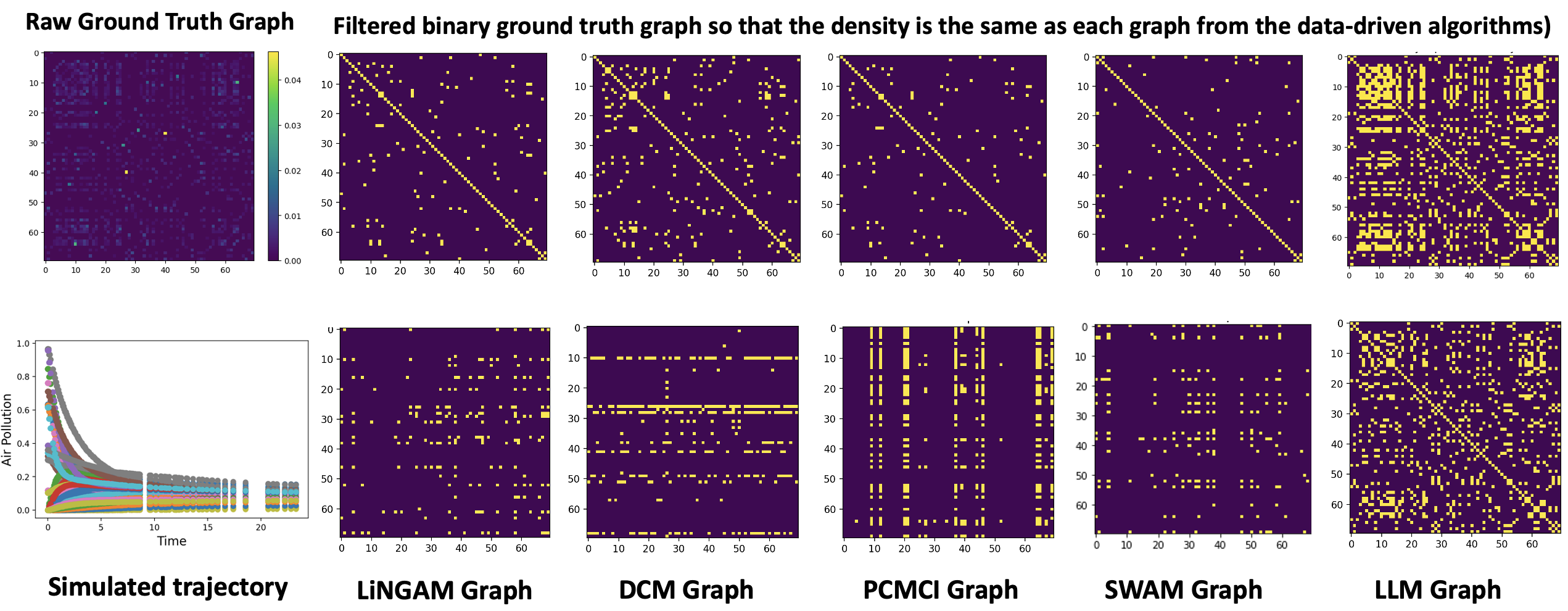}
\caption[graph comparisons]{\textbf{Synthetic Data Experiments for graph comparison with the ground truth} This figure displays the graphs obtained from different data-driven graph learning methods compared with the filtered ground truth graph at the same density.} 
\label{synthetic}
\end{figure}

To more directly evaluate the graph derived from our proposed algorithm and compare it with more algorithms of learning the graph from time series which are unable to construct disease progression trajectory due to their incapabilities of handling continuous irregular data, we generated synthetic data for the comparison purpose where the ground truth of the graph is known and can be queried from LLM. Specifically, we simulate the air pollution of 70 main cities in China by creating a graph of spatial proximity using the inverse of the geodesic distance calculated from the coordinates of each city. Then we simulate the air pollution by using the one-component diffusion process on the proximity network \cite{Raj2012ADementia}, i.e. assuming that the pollution diffuses from the cities of the high concentration of pollution to the rest of the cities, eventually reaching the status that all the cities have the equivalent concentration with time going by. We apply different graph inference methods to the simulated time series data and compare the obtained graph in Figure \ref{synthetic}. It can be observed that LLM can capture the main patterns of the relations while other methods struggle to capture many existing connections.

\subsection{More results for NGM modelling}
\label{Appendix-NGM}

As defined in \cite{bellot2021neural,zou2006adaptive,zhao2006model}, the definition of the GL and AGL regularization are:
\begin{equation}
\rho_{\mathrm{GL}}\left(\mathbf{f}_\theta\right):=\lambda_{\mathrm{GL}} \sum_{k, j=1}^d\left\|\left[A_1^j\right]_{\cdot k}\right\|_2, \quad \rho_{\mathrm{AGL}}\left(\mathbf{f}_\theta\right):=\lambda_{\mathrm{AGL}} \sum_{k, j=1}^d \frac{1}{\left\|\left[\hat{A}_1^j\right]_{\cdot k}\right\|_2^\gamma}\left\|\left[A_1^j\right]_{\cdot k}\right\|_2
\end{equation}

where $\hat{A}^{j}_{i}$ is the GL estimate. The parameters $\lambda_{GL}$ and $\lambda_{AGL}$ control the regularization intensity. Additionally, $\gamma > 0$ and $\|\cdot\|_{2}$ represent the Euclidean norm. \textbf{AGL} utilizes its base estimator to provide a preliminary, data-driven estimate, allowing it to shrink groups of parameters with different regularization strengths. 

\begin{figure}%[H]
\centering
\includegraphics[scale = 0.4]{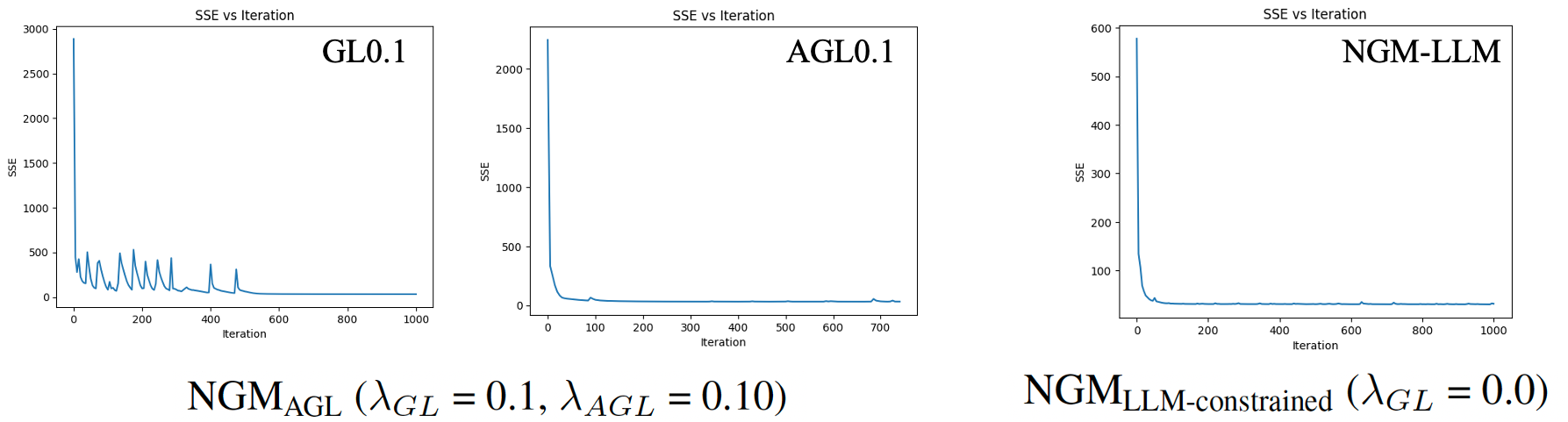}
\caption[Convergence Plot - Comparison between the Adaptive Group Lasso regularization (AGL) and the proposed LLM-constrained regularization]{\textbf{Convergence Plot - Comparison for the AGL regularization and the proposed LLM-constrained regularization} } 
\label{converge_compare}
\end{figure}

Figure \ref{converge_compare} compares the converge plots of the AGL-constrained NGM and the proposed LLM graph-constrained NGM by displaying the SSE vs the number of iterations. Since the formulation of the regularization of AGL is dependent on the weight from GL, the total number of iterations needs to be accumulated. The plots on the left demonstrate that the convergence of the GL method is not stable and needs a relatively large number of iterations to be converged namely 1000 runs. Then followed by the AGL method starting from an initial SSE of around 2000, whose convergence stabilizes after 300 iterations with some vibration afterwards. While for the proposed LLM constrained regularization, converge is achieved around 150 iterations in total starting from an initial SSE of around 600. This shows that the proposed LLM graph-constrained method provides a good regularization from the expert knowledge.

\subsection{Disease Progression Visualization}

\label{brain_map}
Below, we display the disease progression pattern of tau from the real observations vs the fitting using the best pathophysiological model guided by the mixture of LLMs constructed using our proposed framework via brain mapping relative to the pseudo time axis. After allocating all the subjects on the pseudo-time axis, the observations and the model fitting at the relative locations of 0, 1/4, 2/4, 3/4, 4/4 are visualized (if there is no observation at the exact point, the closest observation nearby is taken, and the modelling fitting is taken at the same nearby time). The colour bar, shared by all brain plots, displays the level of normalized tau SUVR at each brain region. It can be observed that the major patterns of tau distribution with time have been captured using our proposed model.

\begin{figure}[H]
\centering
\includegraphics[scale = 0.32]{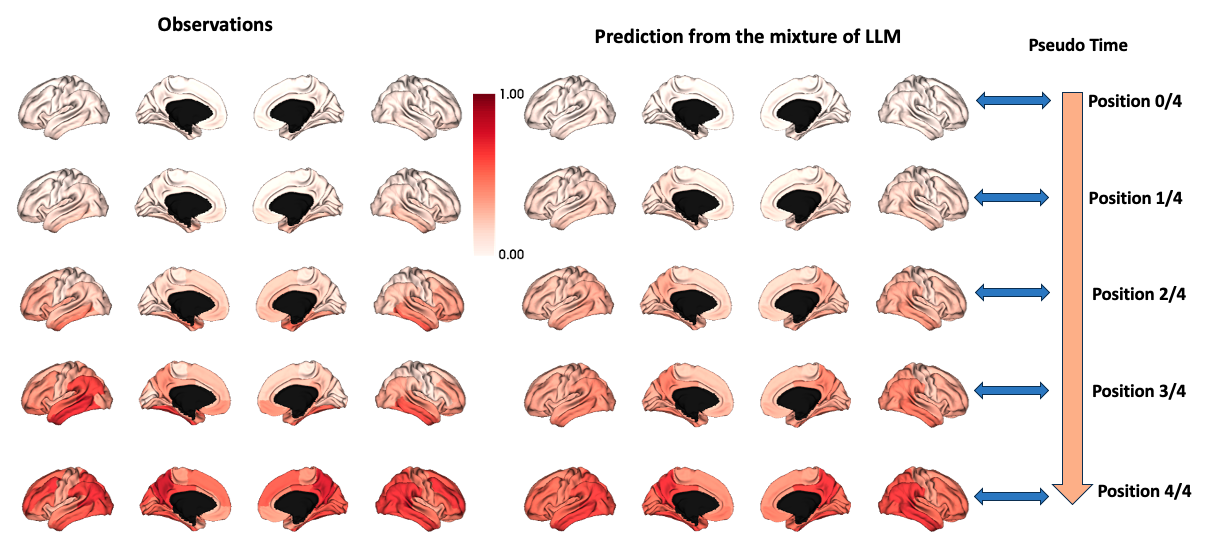}
\caption[The plot displays the tau distribution pattern during disease progression from the real observations vs the fitting using the best pathophysiological model guided by the mixture of LLMs constructed using our proposed framework via brain mapping regarding the pseudo time axis]{The plot displays the tau distribution pattern during disease progression from the real observations vs the fitting using the best pathophysiological model guided by the mixture of LLMs constructed using our proposed framework via brain mapping relative to the pseudo time axis. The colour bar, shared by all brain plots, displays the level of normalized tau SUVR at each brain region.}
\label{explains}
\end{figure}

\subsection{More Interpretation from LLMs}
 Below, we display the reasoning of the same region from the rest of the LLMs, where it can be observed that GPT4-turbo provides the least analysis. This might be one of the reasons that GPT4-turbo performs the worst among the three language models.
\begin{figure}%[H]
\centering
\includegraphics[scale = 0.23]{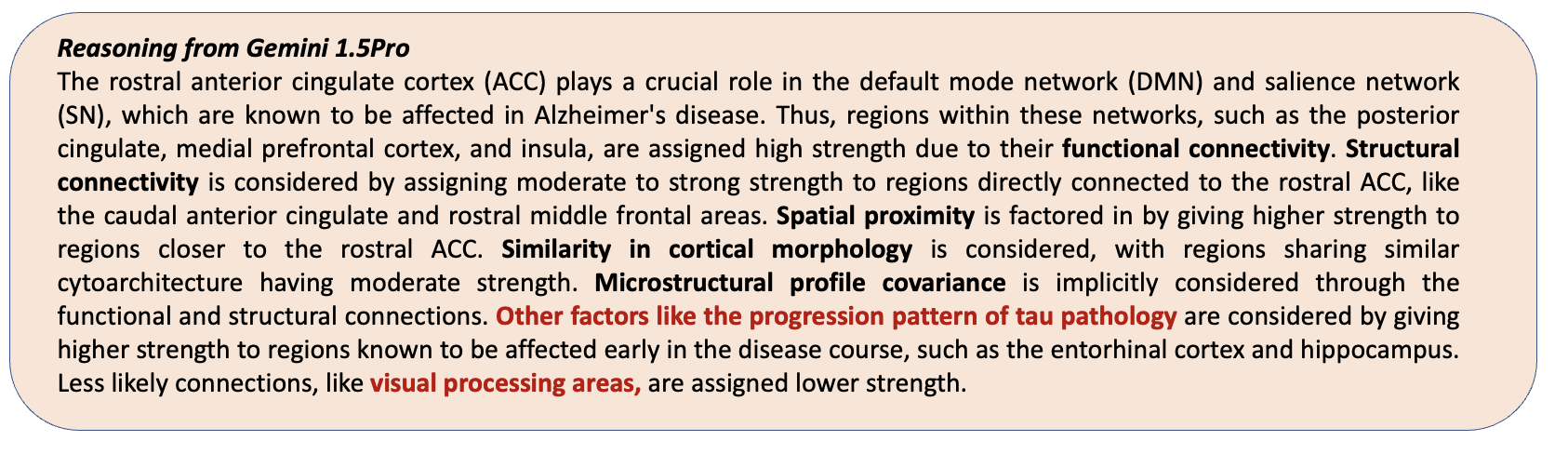}
\caption[One example of reply from Gemini 1.5 Pro]{This figure displays one representative example of an output from Gemini 1.5 Pro.} 
\label{explains}
\end{figure}

\begin{figure}%[H]
\centering
\includegraphics[scale = 0.23]{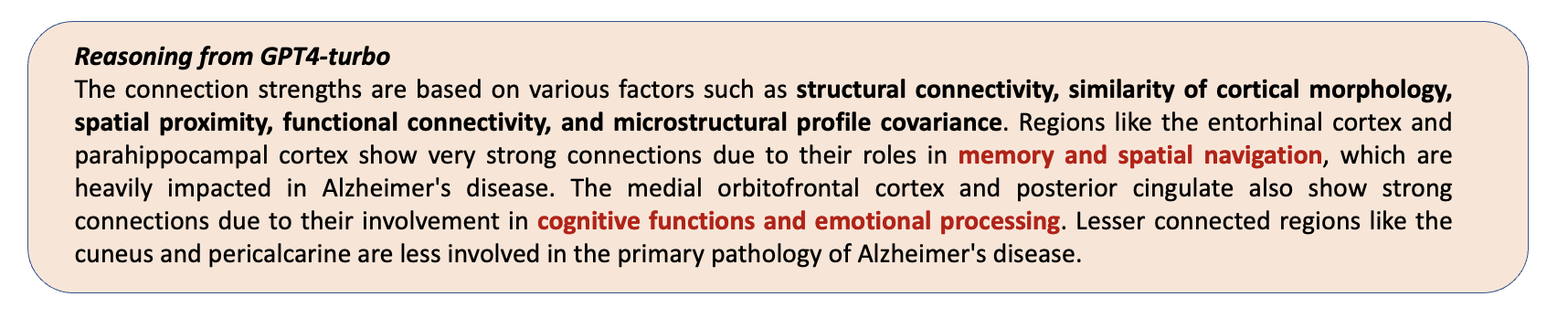}
\caption[One example of reply from GPT4 Turbo]{This figure displays one representative example of an output from GPT4 Turbo.} 
\label{explains}
\end{figure}

\subsubsection{LLM for graph learning}
% \textcolor{red}{see Causal Reasoning and Large Language Models: Opening a New Frontier for Causality Chap3, especially Chap 3.2 Full Graph Discovery}

% \textcolor{red}{And the introduction and related work parts in Causal Modelling Agents: graph Discovery through Synergising Metadata- and Data-driven Reasoning}

Graph learning from time series, the task of uncovering the underlying variable-dependent relationships within a system, plays a critical role in various scientific fields \cite{peters2017elements,glymour2019review}. Causal graph discovery can be one typical representative of uncovering how the variables interact with each other. It often focuses on constructing DAGs, where edges represent causal influences between variables. However, a significant challenge in graph learning lies in identifying the unique true variable-dependent structure. Multiple DAGs can explain the observed data equally well, leading to the issue of non-identifiability \cite{pearl2009causality}. While advancements such as restricting the data-generating process or employing deep learning for modelling variable covariances have been made, pinpointing the single correct graph solely from observational data remains an unsolved problem in many scenarios \cite{kiciman2023causal}.

LLMs offer a promising perspective for addressing the challenges of graph learning by focusing on metadata associated with variables rather than their raw data values. By utilizing the contextual information embedded in variable names and problem domains, LLMs can infer graphs like human domain experts, based on general and domain-specific knowledge. Studies have explored the potential of LLMs for graph learning. %Tu et al. \cite{tu2023causal} applied LLMs directly to a neuropathic pain dataset \cite{tu2019neuropathic}. Kıcıman et al. \cite{kiciman2023causal} investigated the graph discovery capabilities of LLMs across diverse, complex real-world datasets requiring specialized knowledge (e.g., pain diagnosis and Arctic sea ice coverage). They also evaluated the robustness of LLM-based discovery and found that LLMs can achieve good accuracy with appropriate prompting. However, Kıcıman et al. \cite{kiciman2023causal} did not explore combining LLMs with existing causal methods. 
%Since most datasets provide both data and metadata, a promising avenue lies in integrating LLMs with established graph learning methods. LLMs can serve as priors, post-processors, or critics during the learning process, enabling the extraction of variable dependent relationships from both data covariance and metadata. 
Choi et al. \cite{choi2022lmpriors} demonstrated that LLM-generated prior hypotheses can enhance the accuracy of data-driven graph learning algorithms. Long et al. \cite{long2023can} focused on LLMs as a post-processing step, showing their ability to reduce the size of a Markov equivalence class under the assumption of an optimal discovery algorithm output. Abdulaal et al. \cite{abdulaal2023causal} proposed the CMA framework, which synergizes the metadata-based reasoning capabilities of LLMs with the data-driven modeling of DSCMs for graph learning.

\subsubsection{Graph learning for time series}

Graph learning approaches in multivariate time series aim to uncover the causal relationship between time series. Such methods fall into several categories, including well-established approaches like Granger causality, alongside newer methods like constraint-based, score-based, and functional causal model-based approaches\cite{gong2023causal,assaad2022survey}. 

Granger causality is one of the oldest tools for analyzing time series data and inferring potential variable-dependent relationships \cite{granger1969investigating}, forming the foundation for many modern methods. Earlier methods typically use the popular vector autoregressive (VAR) model under the assumption of linear time-series dynamics. However, real-world scenarios often involve non-linear dynamics, particularly in fields like neuroscience or finance \cite{shojaie2022granger}. To address nonlinear dependencies, model-free methods like transfer entropy \cite{vicente2011transfer} and directed information \cite{amblard2011directed} offer an alternative, but they often require substantial data and struggle with high-dimensional settings. Beyond traditional and model-free approaches, researchers have explored other techniques to capture non-linear relationships in time series data. Differential equations excel at capturing non-linear relationships, making them valuable for describing interactions in dynamic systems. Recent work proposes Neural Graphical Models (NGMs), which model the latent vector field explicitly with penalized extensions to Neural ODEs \cite{bellot2021neural}. Other neural networks like MLP, RNN, LSTM can also be combined with Granger causality methods for modelling the complex and non-linear dynamics \cite{gong2023causal,shojaie2022granger}.

Another powerful tool for uncovering variable dependent relationships, constraint-based discovery methods work in two stages. First, it uses statistical tests to identify potential connections between variables, building a network of possible links. Then, specific rules are applied to orient these connections, resulting in a directed acyclic graph (DAG) that reflects the most basic causal structure between the variables. These approaches often rely on assumptions like the causal Markov property and faithfulness \cite{gong2023causal}. A prominent example is the Peter-Clark (PC) algorithm, which streamlines the process by reducing unnecessary tests, specifically for non-temporal data with the assumption of causal sufficiency. To handle time series data, the PC algorithm has been extended with methods like optimal causation entropy (oCSE) \cite{sun2015causal}, which leverages transfer entropy, and PCMCI \cite{runge2020discovering} which uses momentary conditional independence tests.

Functional Causal\ Models (FCMs), also known as Structural Equation Models (SEMs) \cite{neuberg2003causality}, describe a causal system using a set of equations. Each equation explains how a variable depends on its direct causes and an error term. This allows FCMs to capture both linear and non-linear relationships between variables. VAR-LiNGAM \cite{hyvarinen2008causal,hyvarinen2010estimation}, a typical FCM-based graph learning algorithm for time series, is built upon the non-temporal LiNGAM model \cite{shimizu2006linear} and estimates structural autoregressive (SVAR) models by exploiting non-Gaussianity properties in the data. Another family of FCMs is based on the additive noise model (ANM), offering more flexibility by incorporating non-linear functions within its framework. It relaxes the linear constraints of VAR-LiNGAM and is suitable for more complex scenarios. An example of this family is the Time Series Models with Independent Noise (TiMINo) method \cite{peters2013causal}.

In score-based approaches, a graph corresponds to a probabilistic (or Bayesian) network;
furthermore, a dynamic probabilistic (or dynamic Bayesian) network (DPN) is a probabilistic network in which variables are time series \cite{assaad2022survey}. score-based methods aim at finding sparse structural equation models that best
explain the data, without any guarantee on the corresponding DAG (Kaiser and Sipos, 2021). This
contrasts with, e.g., constraint-based approaches. 

Score-based graph learning methods view variable-dependent relationships as a Bayesian network or a dynamic Bayesian network dealing with temporal data \cite{assaad2022survey}. Score-based methods prioritize finding a simple model that best explains the data, even if it does not perfectly map out the exact causal structure (DAG). This is in contrast to constraint-based methods, which focus on precisely identifying those causal connections. Friedman et al. \cite{friedman2013learning} first use the Structural Expectation-Maximization (Structural EM) algorithm \cite{friedman1997learning,friedman1998bayesian} to infer a Dynamic Bayesian Network (DBN) from longitudinal data. Pamfil et al. \cite{pamfil2020dynotears} proposed DYNOTEARS, a method that can simultaneously capture contemporaneous and time-lagged relationships between time series. To overcome the limitation of DYNOTEARS, which is a linear autoregressive model, NTS-NOTEARS \cite{sun2021nts} is proposed based on 1D CNNs to extract both linear and non-linear relations among variables.

\end{document}

%% file: math_commands.tex
\usepackage{amsmath,amsfonts,bm}

\def\eqref#1{equation~\ref{#1}}
\def\1{\bm{1}}

\DeclareMathAlphabet{\mathsfit}{\encodingdefault}{\sfdefault}{m}{sl}
\SetMathAlphabet{\mathsfit}{bold}{\encodingdefault}{\sfdefault}{bx}{n}

%% file: arxiv.bib
@article{he2025stage,
  title={A Stage-Aware Mixture of Experts Framework for Neurodegenerative Disease Progression Modelling},
  author={He, Tiantian and Jiang, Keyue and Zhao, An and Schroder, Anna and Thompson, Elinor and Soskic, Sonja and Barkhof, Frederik and Alexander, Daniel C},
  journal={arXiv preprint arXiv:2508.07032},
  year={2025}
}

@inproceedings{soskic2024effects,
  title={Effects of regional neurotransmitter receptor densities on modelling amyloid and tau accumulation in Alzheimer’s disease with network spreading models},
  author={Soskic, Sonja and Thompson, Elinor and He, Tiantian and Schroder, Anna and Oxtoby, Neil P and Alexander, Daniel C},
  booktitle={Alzheimer's Association International Conference},
  year={2024},
  organization={ALZ}
}

@inproceedings{he2023coupled,
  title={A coupled-mechanisms modelling framework for neurodegeneration},
  author={He, Tiantian and Thompson, Elinor and Schroder, Anna and Oxtoby, Neil P and Abdulaal, Ahmed and Barkhof, Frederik and Alexander, Daniel C},
  booktitle={International Conference on Medical Image Computing and Computer-Assisted Intervention},
  pages={459--469},
  year={2023},
  organization={Springer}
}

@misc{Young2024Data-drivenBox,
    title = {{Data-driven modelling of neurodegenerative disease progression: thinking outside the black box}},
    year = {2024},
    booktitle = {Nature Reviews Neuroscience},
    author = {Young, Alexandra L. and Oxtoby, Neil P. and Garbarino, Sara and Fox, Nick C. and Barkhof, Frederik and Schott, Jonathan M. and Alexander, Daniel C.},
    publisher = {Springer Nature},
    doi = {10.1038/s41583-023-00779-6},
    issn = {14710048}
}

@article{Raj2012ADementia,
    title = {{A Network Diffusion Model of Disease Progression in Dementia}},
    year = {2012},
    journal = {Neuron},
    author = {Raj, Ashish and Kuceyeski, Amy and Weiner, Michael},
    number = {6},
    month = {3},
    pages = {1204--1215},
    volume = {73},
    doi = {10.1016/j.neuron.2011.12.040},
    issn = {08966273},
    pmid = {22445347}
}

@article{Busche2020SynergyDisease,
    title = {{Synergy between amyloid-{$\beta$} and tau in Alzheimer’s disease}},
    year = {2020},
    journal = {Nature neuroscience},
    author = {Busche, Marc Aurel and Hyman, Bradley T},
    number = {10},
    pages = {1183--1193},
    volume = {23},
    publisher = {Nature Publishing Group US New York},
    issn = {1097-6256}
}

@article{seguin2023brain,
  title={Brain network communication: concepts, models and applications},
  author={Seguin, Caio and Sporns, Olaf and Zalesky, Andrew},
  journal={Nature reviews neuroscience},
  volume={24},
  number={9},
  pages={557--574},
  year={2023},
  publisher={Nature Publishing Group UK London}
}

@article{bellot2021neural,
  title={Neural graphical modelling in continuous-time: consistency guarantees and algorithms},
  author={Bellot, Alexis and Branson, Kim and van der Schaar, Mihaela},
  journal={arXiv preprint arXiv:2105.02522},
  year={2021}
}

@article{kiciman2023causal,
  title={Causal reasoning and large language models: Opening a new frontier for causality},
  author={K{\i}c{\i}man, Emre and Ness, Robert and Sharma, Amit and Tan, Chenhao},
  journal={arXiv preprint arXiv:2305.00050},
  year={2023}
}

@inproceedings{abdulaal2023causal,
  title={Causal Modelling Agents: Causal Graph Discovery through Synergising Metadata-and Data-driven Reasoning},
  author={Abdulaal, Ahmed and Montana-Brown, Nina and He, Tiantian and Ijishakin, Ayodeji and Drobnjak, Ivana and Castro, Daniel C and Alexander, Daniel C and others},
  booktitle={The Twelfth International Conference on Learning Representations},
  year={2023}
}

@article{choi2022lmpriors,
  title={Lmpriors: Pre-trained language models as task-specific priors},
  author={Choi, Kristy and Cundy, Chris and Srivastava, Sanjari and Ermon, Stefano},
  journal={arXiv preprint arXiv:2210.12530},
  year={2022}
}

@article{long2023can,
  title={Can large language models build causal graphs?},
  author={Long, Stephanie and Schuster, Tibor and Pich{\'e}, Alexandre and de Montreal, Universit{\'e} and Research, ServiceNow and others},
  journal={arXiv preprint arXiv:2303.05279},
  year={2023}
}

@article{gong2023causal,
  title={Causal discovery from temporal data: An overview and new perspectives},
  author={Gong, Chang and Yao, Di and Zhang, Chuzhe and Li, Wenbin and Bi, Jingping},
  journal={arXiv preprint arXiv:2303.10112},
  year={2023}
}

@article{assaad2022survey,
  title={Survey and evaluation of causal discovery methods for time series},
  author={Assaad, Charles K and Devijver, Emilie and Gaussier, Eric},
  journal={Journal of Artificial Intelligence Research},
  volume={73},
  pages={767--819},
  year={2022}
}

@article{granger1969investigating,
  title={Investigating causal relations by econometric models and cross-spectral methods},
  author={Granger, Clive WJ},
  journal={Econometrica: journal of the Econometric Society},
  pages={424--438},
  year={1969},
  publisher={JSTOR}
}

@article{shojaie2022granger,
  title={Granger causality: A review and recent advances},
  author={Shojaie, Ali and Fox, Emily B},
  journal={Annual Review of Statistics and Its Application},
  volume={9},
  pages={289--319},
  year={2022},
  publisher={Annual Reviews}
}

@article{vicente2011transfer,
  title={Transfer entropy—a model-free measure of effective connectivity for the neurosciences},
  author={Vicente, Raul and Wibral, Michael and Lindner, Michael and Pipa, Gordon},
  journal={Journal of computational neuroscience},
  volume={30},
  number={1},
  pages={45--67},
  year={2011},
  publisher={Springer}
}

@article{amblard2011directed,
  title={On directed information theory and Granger causality graphs},
  author={Amblard, Pierre-Olivier and Michel, Olivier JJ},
  journal={Journal of computational neuroscience},
  volume={30},
  number={1},
  pages={7--16},
  year={2011},
  publisher={Springer}
}

@article{sun2015causal,
  title={Causal network inference by optimal causation entropy},
  author={Sun, Jie and Taylor, Dane and Bollt, Erik M},
  journal={SIAM Journal on Applied Dynamical Systems},
  volume={14},
  number={1},
  pages={73--106},
  year={2015},
  publisher={SIAM}
}

@inproceedings{runge2020discovering,
  title={Discovering contemporaneous and lagged causal relations in autocorrelated nonlinear time series datasets},
  author={Runge, Jakob},
  booktitle={Conference on Uncertainty in Artificial Intelligence},
  pages={1388--1397},
  year={2020},
  organization={Pmlr}
}

@article{neuberg2003causality,
  title={Causality: models, reasoning, and inference, by judea pearl, cambridge university press, 2000},
  author={Neuberg, Leland Gerson},
  journal={Econometric Theory},
  volume={19},
  number={4},
  pages={675--685},
  year={2003},
  publisher={cambridge university press}
}

@inproceedings{hyvarinen2008causal,
  title={Causal modelling combining instantaneous and lagged effects: an identifiable model based on non-Gaussianity},
  author={Hyv{\"a}rinen, Aapo and Shimizu, Shohei and Hoyer, Patrik O},
  booktitle={Proceedings of the 25th international conference on Machine learning},
  pages={424--431},
  year={2008}
}

@article{hyvarinen2010estimation,
  title={Estimation of a structural vector autoregression model using non-gaussianity.},
  author={Hyv{\"a}rinen, Aapo and Zhang, Kun and Shimizu, Shohei and Hoyer, Patrik O},
  journal={Journal of Machine Learning Research},
  volume={11},
  number={5},
  year={2010}
}

@article{shimizu2006linear,
  title={A linear non-Gaussian acyclic model for causal discovery.},
  author={Shimizu, Shohei and Hoyer, Patrik O and Hyv{\"a}rinen, Aapo and Kerminen, Antti and Jordan, Michael},
  journal={Journal of Machine Learning Research},
  volume={7},
  number={10},
  year={2006}
}

@article{peters2013causal,
  title={Causal inference on time series using restricted structural equation models},
  author={Peters, Jonas and Janzing, Dominik and Sch{\"o}lkopf, Bernhard},
  journal={Advances in neural information processing systems},
  volume={26},
  year={2013}
}

@article{friedman2013learning,
  title={Learning the structure of dynamic probabilistic networks},
  author={Friedman, Nir and Murphy, Kevin and Russell, Stuart},
  journal={arXiv preprint arXiv:1301.7374},
  year={2013}
}

@inproceedings{friedman1997learning,
  title={Learning belief networks in the presence of missing values and hidden variables},
  author={Friedman, Nir and others},
  booktitle={Icml},
  volume={97},
  number={July},
  pages={125--133},
  year={1997},
  organization={Berkeley, CA}
}

@inproceedings{friedman1998bayesian,
  title={The Bayesian structural EM algorithm},
  author={FRIEDMAN, N},
  booktitle={Proc. Conf. on Uncertainty in Artificial Intelligence (UAI-98)},
  pages={129--138},
  year={1998}
}

@inproceedings{pamfil2020dynotears,
  title={Dynotears: Structure learning from time-series data},
  author={Pamfil, Roxana and Sriwattanaworachai, Nisara and Desai, Shaan and Pilgerstorfer, Philip and Georgatzis, Konstantinos and Beaumont, Paul and Aragam, Bryon},
  booktitle={International Conference on Artificial Intelligence and Statistics},
  pages={1595--1605},
  year={2020},
  organization={Pmlr}
}

@article{sun2021nts,
  title={NTS-NOTEARS: Learning nonparametric DBNs with prior knowledge},
  author={Sun, Xiangyu and Schulte, Oliver and Liu, Guiliang and Poupart, Pascal},
  journal={arXiv preprint arXiv:2109.04286},
  year={2021}
}

@book{peters2017elements,
  title={Elements of causal inference: foundations and learning algorithms},
  author={Peters, Jonas and Janzing, Dominik and Sch{\"o}lkopf, Bernhard},
  year={2017},
  publisher={The MIT Press}
}

@article{glymour2019review,
  title={Review of causal discovery methods based on graphical models},
  author={Glymour, Clark and Zhang, Kun and Spirtes, Peter},
  journal={Frontiers in genetics},
  volume={10},
  pages={524},
  year={2019},
  publisher={Frontiers Media SA}
}

@book{pearl2009causality,
  title={Causality},
  author={Pearl, Judea},
  year={2009},
  publisher={Cambridge university press}
}

@article{raj2012network,
  title={A network diffusion model of disease progression in dementia},
  author={Raj, Ashish and Kuceyeski, Amy and Weiner, Michael},
  journal={Neuron},
  volume={73},
  number={6},
  pages={1204--1215},
  year={2012},
  publisher={Elsevier}
}

@article{garbarino2019differences,
  title={Differences in topological progression profile among neurodegenerative diseases from imaging data},
  author={Garbarino, Sara and Lorenzi, Marco and Oxtoby, Neil P and Vinke, Elisabeth J and Marinescu, Razvan V and Eshaghi, Arman and Ikram, M Arfan and Niessen, Wiro J and Ciccarelli, Olga and Barkhof, Frederik and others},
  journal={Elife},
  volume={8},
  pages={e49298},
  year={2019},
  publisher={eLife Sciences Publications, Ltd}
}

@article{weickenmeier2018multiphysics,
  title={Multiphysics of prionlike diseases: Progression and atrophy},
  author={Weickenmeier, Johannes and Kuhl, Ellen and Goriely, Alain},
  journal={Physical review letters},
  volume={121},
  number={15},
  pages={158101},
  year={2018},
  publisher={APS}
}

@article{thompson2024combining,
  title={Combining multimodal connectivity information improves modelling of pathology spread in Alzheimer’s disease},
  author={Thompson, Elinor and Schroder, Anna and He, Tiantian and Shand, Cameron and Soskic, Sonja and Oxtoby, Neil P and Barkhof, Frederik and Alexander, Daniel C and Alzheimer’s Disease Neuroimaging Initiative},
  journal={Imaging Neuroscience},
  volume={2},
  pages={1--19},
  year={2024},
  publisher={MIT Press One Broadway, 12th Floor, Cambridge, Massachusetts 02142, USA~…}
}

@article{zhou2012predicting,
  title={Predicting regional neurodegeneration from the healthy brain functional connectome},
  author={Zhou, Juan and Gennatas, Efstathios D and Kramer, Joel H and Miller, Bruce L and Seeley, William W},
  journal={Neuron},
  volume={73},
  number={6},
  pages={1216--1227},
  year={2012},
  publisher={Elsevier}
}

@misc{Vogel2023Connectome-basedInsight,
    title = {{Connectome-based modelling of neurodegenerative diseases: towards precision medicine and mechanistic insight}},
    year = {2023},
    booktitle = {Nature Reviews Neuroscience},
    author = {Vogel, Jacob W. and Corriveau-Lecavalier, Nick and Franzmeier, Nicolai and Pereira, Joana B. and Brown, Jesse A. and Maass, Anne and Botha, Hugo and Seeley, William W. and Bassett, Dani S. and Jones, David T. and Ewers, Michael},
    number = {10},
    month = {10},
    pages = {620--639},
    volume = {24},
    publisher = {Springer Nature},
    doi = {10.1038/s41583-023-00731-8},
    issn = {14710048},
    pmid = {37620599}
}

@article{zhao2006model,
  title={On model selection consistency of Lasso},
  author={Zhao, Peng and Yu, Bin},
  journal={The Journal of Machine Learning Research},
  volume={7},
  pages={2541--2563},
  year={2006},
  publisher={JMLR. org}
}

@article{zou2006adaptive,
  title={The adaptive lasso and its oracle properties},
  author={Zou, Hui},
  journal={Journal of the American statistical association},
  volume={101},
  number={476},
  pages={1418--1429},
  year={2006},
  publisher={Taylor \& Francis}
}

@article{adams2019relationships,
  title={Relationships between tau and glucose metabolism reflect Alzheimer’s disease pathology in cognitively normal older adults},
  author={Adams, Jenna N and Lockhart, Samuel N and Li, Lexin and Jagust, William J},
  journal={Cerebral cortex},
  volume={29},
  number={5},
  pages={1997--2009},
  year={2019},
  publisher={Oxford University Press}
}
